\documentclass{article}

\usepackage[preprint, nonatbib]{neurips_2024}
\usepackage[numbers]{natbib} 

\usepackage[utf8]{inputenc} 
\usepackage[T1]{fontenc}    
\usepackage{url}            
\usepackage{booktabs}       
\usepackage{amsfonts}       
\usepackage{nicefrac}       
\usepackage{microtype}      
\usepackage{xcolor}         
\usepackage{amsmath}
\usepackage{todonotes}
\usepackage{tikz}
\usepackage{makecell}

\title{Decoupling Knowledge and Reasoning in Transformers: A Modular Architecture with Generalized Cross-Attention}

\author{
Zhenyu Guo$^{1,2}$\\
$^1$ Tsinghua University\\
$^2$ Ant Group\\
\texttt{imzhenyu@outlook.com}
\And
Wenguang Chen\\
Tsinghua University\\
\texttt{cwg@tsinghua.edu.cn}
}

\begin{document}

\maketitle

\begin{abstract}
    Transformers have achieved remarkable success across diverse domains, but their monolithic architecture presents challenges in interpretability, adaptability, and scalability. This paper introduces a novel modular Transformer architecture that explicitly decouples knowledge and reasoning through a generalized cross-attention mechanism to a globally shared knowledge base with layer-specific transformations, specifically designed for effective knowledge retrieval. Critically, we provide a rigorous mathematical derivation demonstrating that the Feed-Forward Network (FFN) in a standard Transformer is a specialized case (a closure) of this generalized cross-attention, revealing its role in implicit knowledge retrieval and validating our design. This theoretical framework provides a new lens for understanding FFNs and lays the foundation for future research exploring enhanced interpretability, adaptability, and scalability, enabling richer interplay with external knowledge bases and other systems.
\end{abstract}

\section{Introduction}\label{sec:intro}

Large language models (LLMs) based on the Transformer architecture have achieved remarkable success~\cite{vaswani2017attention,devlin2018bert,radford2018improving,brown2020language}. However, a key limitation of their monolithic design is the deep entanglement of knowledge and reasoning within the model's parameters, posing significant challenges for real-world applications requiring transparency, adaptability, and continuous learning.

Specifically, current Transformers face limitations in:

\begin{itemize}
    \item \textbf{Interpretability.} The distributed nature of knowledge representation, especially within Feed-Forward Networks (FFNs), makes it difficult to pinpoint the specific information used for predictions~\cite{clark2019bert,vig2019multiscale}, limiting their use in high-stakes applications like healthcare and legal reasoning.
    \item \textbf{Adaptability.} Adapting pre-trained models to new knowledge or integrating them with external systems is inefficient and complex. Current methods like RAG~\cite{lewis2020retrieval}, which simply concatenate retrieved context with input, suffer from context dilution and retrieval performed only at the input level, lacking dynamic, context-dependent access to knowledge during processing. This hinders continuous learning and the ability to incorporate rapidly changing information, such as real-time news or scientific discoveries.
    \item \textbf{Scalability.} The tight coupling of knowledge and reasoning in monolithic Transformers prevents \textit{modular} scaling, unlike human cognition, which efficiently transfers existing reasoning capabilities to new knowledge through modular learning. This "parameter explosion" limits accessibility of larger models, hindering progress towards truly comprehensive knowledge-driven AI.
\end{itemize}

To address these limitations, while the interpretation of FFNs as implicit key-value memories~\cite{geva2020transformer} has provided valuable insights, a new challenge emerges when we seek to explicitly decouple knowledge and reasoning: how to maintain global knowledge consistency while enabling contextualized access at each layer. Current approaches, including those derived from the key-value perspective, often struggle to reconcile these two competing demands. To address this gap, we propose a novel modular Transformer architecture that explicitly decouples knowledge and reasoning through a generalized cross-attention mechanism to a globally shared knowledge base with layer-specific transformations. Our core goal is to enable seamless interaction with external knowledge bases, facilitating continuous learning, knowledge sharing, and independent scaling of knowledge capacity.

A key contribution of this work is a rigorous theoretical analysis demonstrating that the FFN in a standard Transformer can be expressed as a specialized case (a closure) of our generalized cross-attention. This reveals the implicit knowledge retrieval role of FFNs and provides a crucial validation of our proposed mechanism. By establishing this functional equivalence under joint training (where the knowledge base is trained within the model), we provide a solid theoretical foundation for future exploration of external knowledge base integration. Due to this proven equivalence, we expect functionally identical performance in this joint training setting. This theoretical groundwork is essential for the subsequent exploration of external knowledge integration, which is the primary focus of future work.

This paper makes the following key contributions:

\begin{itemize}
    \item \textbf{A Novel Modular Architecture.} We propose a modular Transformer architecture that explicitly decouples knowledge and reasoning through generalized cross-attention to a shared knowledge base with layer-specifc transformations.
    \item \textbf{Generalized Cross-Attention for Knowledge Retrieval.} We introduce a generalized cross-attention mechanism specifically designed for effective knowledge retrieval, incorporating knowledge-specific biases.
    \item \textbf{Novel FFN Interpretation and Framework Validation.} We provide a rigorous mathematical derivation demonstrating that the FFN in a standard Transformer is a specialized case of our generalized cross-attention, revealing their role in implicit knowledge retrieval and validating our design.
    \item \textbf{A Foundation for Enhanced Capabilities.} This theoretical framework lays the foundation for future research exploring enhanced interpretability, adaptability, and scalability, enabling richer interplay with external knowledge bases and other systems.
\end{itemize}

This paper focuses on the \textit{theoretical framework} and the case where the shared knowledge base is trained \textit{jointly within the model}, laying the groundwork for future exploration of external, pluggable knowledge bases and associated empirical evaluations.

The rest of the paper is organized as follows: Section~\ref{sec:challenges} details the challenges of monolithic Transformers. Section~\ref{sec:modular} presents our proposed modular architecture. Section~\ref{sec:attn} describes our generalized cross-attention. Section~\ref{sec:ffn} presents the theoretical analysis connecting FFNs to our generalized cross-attention. Section~\ref{sec:discuss} discusses implications and future work. Section~\ref{sec:related} reviews related work, and Section~\ref{sec:cc} concludes the paper.

\section{Challenges of Monolithic Transformers}\label{sec:challenges}

This section outlines the limitations of the standard Transformer architecture, focusing on the conceptual challenges arising from the monolithic nature of these models, particularly concerning the entanglement of knowledge and reasoning.

\subsection{The Decoder-Only Transformer Baseline}

\begin{figure}[t] 
    \centering
    \includegraphics[width=0.8\textwidth, clip]{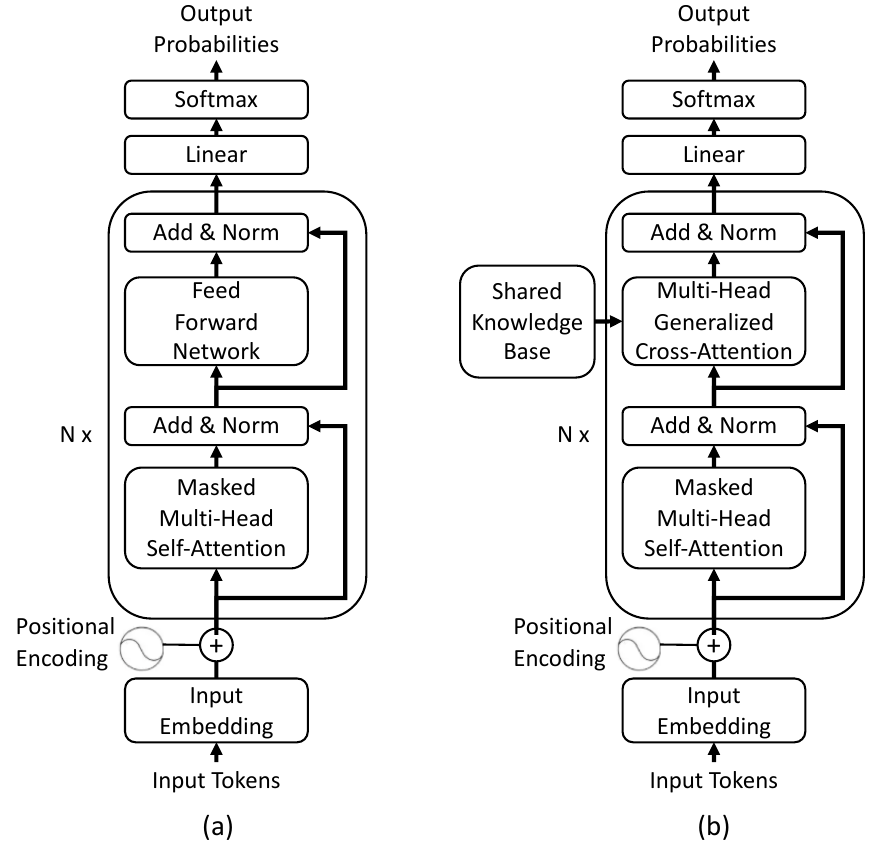} 
    \caption{Architectures for (a) standard decoder-only Transformer, and (b) our proposed modular Transformer~(b) with generalized cross-attention and shared knowledge base.}
    \label{fig:arch} 
\end{figure}

We focus our analysis on the decoder-only Transformer architecture~\cite{vaswani2017attention,radford2018improving,brown2020language}, which has become the dominant architecture for large language models. A decoder-only Transformer, as illustrated in Figure~\ref{fig:arch}~(a), consists of stacked decoder blocks, each containing two main sub-layers:

\begin{itemize}
    \item \textbf{Masked Multi-Head Self-Attention.} This layer allows each token in the input sequence to attend to all preceding tokens (including itself), capturing contextual relationships within the sequence. The "masked" aspect prevents the model from attending to future tokens during training, ensuring autoregressive behavior.
    \item \textbf{Feed-Forward Network (FFN).} This layer consists of two linear transformations with a non-linear activation function (typically GeLU or ReLU) in between. It processes each token's representation independently.
\end{itemize}

The output of each sub-layer is added to the input (residual connection) and normalized (layer normalization).

\subsection{Limitations of the Monolithic Architecture}

The monolithic architecture of standard Transformers, where knowledge and reasoning are deeply intertwined within the model's parameters, presents several key conceptual challenges:

\begin{itemize}
    \item \textbf{Intertwined Knowledge and Reasoning.} In standard Transformers, knowledge is implicitly encoded within the weights of the attention matrices and FFNs. This entanglement entangles knowledge, making it difficult to isolate, analyze, or update specific information without unintended consequences. This lack of transparency makes it hard to determine whether errors stem from a lack of knowledge, flawed reasoning, or complex interactions between the two, hindering targeted improvements. For instance, the FFN for the word "bank" might encode both the concept of a financial institution and the side of a river, making it difficult to disambiguate in different contexts.
    \item \textbf{Adaptability and Non-Modularity.} Adapting pre-trained models to new knowledge or integrating them with external systems is inefficient and complex. Current retrieval-augmented methods like RAG attempt to improve adaptability by retrieving relevant context and concatenating it with the input query. However, this approach has several inherent limitations. First, simply adding more text through concatenation can lead to "context dilution", where the factual knowledge and user question are mixed, confusing the LLM. Second, because retrieval and reasoning remain entangled within the model, it becomes difficult to understand how the retrieved knowledge is actually being used. Finally, due to RAG's single initial retrieval and the limited input window of LLMs, the retrieved information is often insufficient or excessive. In contrast, without any retrieval mechanism like RAG, incorporating new information becomes even more challenging, requiring significant resources and risking disruption to existing knowledge.
    \item \textbf{Scaling Challenges.} The monolithic structure of Transformers presents significant scaling challenges. Because knowledge and reasoning are deeply intertwined within the model's parameters, increasing one capacity necessitates significant adjustments to the other, preventing independent scaling and leading to disproportionate parameter growth and substantial computational and memory costs. This "parameter explosion" makes training and deploying larger models increasingly difficult. While fine-tuning offers a more efficient alternative to full retraining for incorporating new knowledge, it still suffers from limitations such as catastrophic forgetting and the lack of efficient, additive scaling. Unlike humans, who can efficiently integrate new knowledge with minimal adjustments to their core reasoning abilities (achieving a form of \textit{additive} scaling), monolithic Transformers, whether through retraining or fine-tuning, require substantial parameter updates, hindering continuous learning and contributing to diminishing returns: after a certain point, simply increasing model size or performing further fine-tuning yields progressively smaller performance gains, indicating an inefficient approach to scaling knowledge and reasoning.
\end{itemize}
\section{A Modular Architecture with Explicit Knowledge Decoupling}\label{sec:modular}

To address the limitations of monolithic Transformers, particularly the entanglement of knowledge and reasoning (as discussed in Section~\ref{sec:challenges}), we propose a novel modular architecture. This architecture is designed to explicitly decouple knowledge and reasoning by introducing a dedicated mechanism for knowledge access.

\subsection{Requirements for Explicit Knowledge Base}\label{sec:modular:requirements}

To effectively integrate knowledge into Transformer architectures, we believe a valid knowledge base (KB) must satisfy two key requirements:

\begin{itemize}
    \item \textbf{Global sharing.} A globally shared KB ensures \textbf{parameter efficiency} by avoiding redundant storage of the same knowledge across multiple layers. If each layer had its own KB, the number of parameters would increase significantly, leading to larger and more computationally expensive models. Second, a shared KB promotes \textbf{knowledge consistency} by ensuring that all parts of the model are working with the same information, preventing contradictions and inconsistencies. Finally, a centralized KB facilitates \textbf{efficient management} of knowledge, as updating, maintaining, and managing a single KB is much simpler than managing multiple distributed KBs, which is crucial for continuous learning and incorporating new information.
    \item \textbf{Layer-specific views.} Different layers of a Transformer process information at different levels of abstraction. Therefore, each layer needs to access a specific view of the KB that is relevant to its current context. Analogous to how different areas of the human brain process information in distinct ways, requiring different perspectives on the same underlying knowledge, different layers of a Transformer require different perspectives on the shared KB.
\end{itemize}

\subsection{The Proposed Architecture}\label{sec:modular:arch}

To satisfy the requirements outlined above, our modular architecture introduces a globally shared knowledge base \(E \in \mathbb{R}^{|E| \times d_E}\) across all layers with layer-specific transformations, where \(|E|\) is the number of knowledge entries and \(d_E\) is the dimensionality of each entry. \(E\) is accessed via dedicated cross-attention mechanisms (illustrated in Figure~\ref{fig:arch}~(b)). This effectively decomposes the implicit knowledge encoded within FFNs, resulting significant advantages in terms of interpretability, adaptability, and scalability in future work with external knowledge bases. It is important to note that, in this work, we focus on the \textit{theoretical framework} and the case where \(E\) is trained \textit{jointly within the model}. The exploration of external, pluggable knowledge bases is left for future work.

Specifically, similar to the multi-head self-attention layer in a standard Transformer, each attention head within our modular blocks learns distinct projection matrices (\(W_Q^l\), \(W_K^l\), \(W_V^l\)), effectively focusing on different aspects or subsets of information within \(E\). This mechanism enables dynamic, in-context knowledge retrieval, allowing the model to access and integrate relevant information from \(E\) at each layer of processing.

Formally, let \(H_l \in \mathbb{R}^{N \times d}\) be the output of the multi-head self-attention layer in the \(l\)-th block, where \(N\) is the sequence length and \(d\) is the hidden dimension. The knowledge retrieval process in this block is defined as (skipping details for multi-heads for clarity):

\begin{align}
Q_l = H_l W_Q^l \label{eq:ql}\\
K_l = E W_K^l \label{eq:kl}\\
V_l = E W_V^l \label{eq:vl}\\
C_l = \text{GeneralizedAttention}\left(Q_l, K_l, V_l\right) \label{eq:cl}
\end{align}

where \(W_Q^l \in \mathbb{R}^{d \times d_k}\), \(W_K^l \in \mathbb{R}^{d_E \times d_k}\), and \(W_V^l \in \mathbb{R}^{d_E \times d}\) are the query, key, and value projection matrices specific to the \(l\)-th block, respectively. \(E \in \mathbb{R}^{|E| \times d_E}\) represents the knowledge base. Function \texttt{GeneralizedAttention} with its output \(C_l \in \mathbb{R}^{N \times d}\) will be described in detail in Section~\ref{sec:attn}. The output of the \(l\)-th modular block is then:

\begin{equation}
H_l' = H_l + C_l \label{eq:hlprime}
\end{equation}

This modular design, even in the joint-training setting considered in this paper, provides a valuable conceptual framework for understanding knowledge retrieval in Transformers and has significant implications for future work with external knowledge bases.

\begin{itemize}
    \item \textbf{Explicit Knowledge Representation (Joint Training).} Knowledge is now represented in a separate, explicitly accessible module (\(E\)), even though it is trained jointly within the model in this work. This explicit representation provides a clearer conceptual framework for understanding which information is being used for a given prediction.
    \item \textbf{Foundation for Dynamic, In-Context Knowledge Retrieval.} The use of cross-attention to \(E\) provides a mechanism for dynamic, context-dependent retrieval of relevant knowledge at each layer of the network. This contrasts with methods like RAG, where knowledge is retrieved only once at the beginning. This dynamic retrieval is a key aspect of our proposed framework and motivates the generalized cross-attention mechanism described in the next section.
    \item \textbf{Foundation for Independent Knowledge Base Management.} This design provides a foundation for future independent management of the knowledge base. In future work with external knowledge bases, updates to \(E\) (e.g., adding new facts or correcting existing ones) will not require retraining of the projection matrices or the self-attention mechanism.
    \item \textbf{Foundation for Independent Scaling.} While not explored empirically in this paper, this modular design lays the foundation for independent scaling of knowledge and reasoning capacity in future work with external knowledge bases.
\end{itemize}

\section{Generalized Cross-Attention for Knowledge Retrieval}\label{sec:attn}

This section introduces a generalized cross-attention mechanism designed for knowledge retrieval from a knowledge base \(E \in \mathbb{R}^{|E| \times d_E}\), where \(|E|\) is the number of knowledge entries and \(d_E\) is the dimensionality of each entry. Standard attention mechanisms, while effective for self-attention, are suboptimal for knowledge retrieval due to the need for selective access and the connection of distinct embedding spaces. Effective knowledge retrieval requires: (1) determining the relevance of knowledge entries to a given context (Relevance/Selection), and (2) determining how the selected knowledge should be integrated into the model's representation (Integration/Transformation). We progressively build upon the standard attention mechanism to address these aspects.

Let \(H_l \in \mathbb{R}^{N \times d}\) be the input from the previous layer, where \(N\) is the sequence length and \(d\) is the hidden dimension. Let \(Q_l \in \mathbb{R}^{N \times d_k}\), \(K_l \in \mathbb{R}^{|E| \times d_k}\), and \(V_l \in \mathbb{R}^{|E| \times d}\) be the query, key, and value matrices, respectively.

\subsection{Phase 1: Selective Retrieval with Sparse Activation}

Standard attention computes a weighted average of the values based on a softmax over the query-key similarities:

\begin{equation}
    \text{Attention}(Q_l, K_l, V_l) = \text{softmax}\left(\frac{Q_l K_l^T}{\sqrt{d_k}}\right) V_l
\end{equation}

However, \texttt{softmax} assigns non-zero weights to all knowledge entries, hindering selective knowledge retrieval. To enforce sparsity, we replace \texttt{softmax} with a sparse activation function, such as ReLU:

\begin{equation}
    \text{Attention}_{\text{ReLU}}(Q_l, K_l, V_l) = \text{ReLU}\left(\frac{Q_l K_l^T}{\sqrt{d_k}}\right) V_l
\end{equation}

This element-wise application of ReLU on the similarity matrix thresholds values, enforcing sparsity in the attention matrix. Other sparse activations (e.g., Leaky ReLU, Sparsemax) can also be used.

\subsection{Phase 2: Knowledge-Specific Thresholding ("IF" Condition)}

While ReLU introduces sparsity, it applies a uniform threshold of zero to all knowledge entries. This is suboptimal because different entries have varying levels of relevance. We introduce a knowledge-specific thresholding function \(B1^l(E) \in \mathbb{R}^{N \times |E|}\). This can be interpreted as an "IF" condition: IF the relevance score (from \(Q_l K_l^T\)) for a specific knowledge entry exceeds its corresponding threshold from \(B1^l(E)\), THEN the knowledge entry is considered; otherwise, it is filtered out. We currently implement \(B1^l(E)\) using a Multilayer Perceptron (MLP) applied to each knowledge entry embedding.

\begin{equation}
    \text{Attention}_{\text{ReLU+Threshold}}(Q_l, K_l, V_l) = \text{ReLU}\left(\frac{Q_l K_l^T}{\sqrt{d_k}} + B1^l(E)\right) V_l
\end{equation}

\subsection{Phase 3: Transformation Bias for Semantic Bridging}

The value matrix \(V_l\) represents a transformed view of the knowledge entries, acting as the "THEN VALUE" part of the "IF-THEN" logic. This transformation is already handled by \(W_V^l\) in \(V_l = E W_V^l\), which extracts relevant features from the knowledge entries. Furthermore, unlike self-attention where query and value are from the same embedding space, our generalized cross-attention connects distinct embedding spaces: one for \(H_l\) and the other for \(E\). To further bridge this semantic gap by aligning the transformed knowledge representation with the context representation, we introduce a transformation bias \(b2^l \in \mathbb{R}^{d}\) that is added to the weighted values after the thresholding.

\begin{equation}
    \text{GeneralizedAttention}(Q_l, K_l, V_l) = \text{ReLU}\left(\frac{Q_l K_l^T}{\sqrt{d_k}} + B1^l(E)\right) V_l + b2^l \label{eq:generalized_attention}
\end{equation}

In summary, our generalized cross-attention mechanism addresses the requirements of knowledge retrieval by introducing: (1) \texttt{ReLU} for selective retrieval, (2) knowledge-specific thresholding \(B1^l(E)\) as an "IF" condition, and (3) a transformation bias \(b2^l\) for semantic bridging. This design enables more interpretable, effective, and targeted knowledge retrieval compared to standard attention.

\section{FFN is a Closure of Generalized Cross-Attention}\label{sec:ffn}

This section establishes a crucial theoretical link between our proposed modular architecture and the standard Transformer architecture. We demonstrate that the Feed-Forward Network (FFN) within a Transformer block can be interpreted as a specialized case of our generalized cross-attention mechanism.

Consider the generalized cross-attention as a function with two arguments: a query \(H_l\) and a knowledge base \(E\):

\begin{equation}
    \text{Cross-Attention}(H_l, E)
\end{equation}

Our key finding is that the FFN can be expressed as a \textit{closure} of this function:

\begin{equation}
    \text{FFN}(H_l) = \text{Cross-Attention}(H_l, \text{Implicit } E)
\end{equation}

Where "Implicit \(E\)" represents the knowledge encoded within the Transformer's parameters. This 'Implicit E' can be understood as a highly compressed representation of knowledge learned during pre-training, encoded within the weights of the FFN spanning all decoder layers. This representation highlights that the FFN performs implicit knowledge retrieval from a built-in knowledge base. This connection provides a strong theoretical justification for the effectiveness of FFNs and simultaneously validates the design of our generalized cross-attention mechanism. Critically, this derivation provides a formal basis for the key-value memory interpretation of FFNs proposed by Geva et al.~\cite{geva2020transformer}.

Now, we will provide the mathematical derivation that demonstrates this equivalence.

\subsection{Derivation of FFN from Generalized Cross-Attention}\label{ssec:ffn_derivation}

To establish a connection with the standard FFN formulation, which operates on fixed weights, we consider the scenario where \(E\) is static during inference (and, in the joint-training case considered in this paper, static during training as well). In this case, the generalized cross-attention mechanism simplifies significantly. Recall equation~\ref{eq:generalized_attention}:

\begin{equation}
    C_l = \text{ReLU}\left(\frac{Q_l K_l^T}{\sqrt{d_k}} + B1^l(E)\right) V_l + b2^l \label{eq:generalized_al_recalled}
\end{equation}

where:

\begin{align}
Q_l &= H_l W_Q^l \\
K_l &= E W_K^l \\
V_l &= E W_V^l
\end{align}

Because \(E\) is static, we can pre-compute the following matrices, effectively "folding" the implicit knowledge base into the weights:

\begin{align}
W^l_{(K, E)} &= E W_K^l \label{eq:folded_wk}\\
W^l_{(V, E)} &= E W_V^l \label{eq:folded_wv}\\
B1^l_{(E)} &= B1^l(E) \label{eq:folded_b1}
\end{align}

This "folding" process is precisely what creates the closure, making the cross-attention function operate with only the query \(H_l\) as an explicit argument. Substituting these pre-computed terms into Equation \ref{eq:generalized_al_recalled}, we get:

\begin{equation}
    C_l = \text{ReLU}\left(\frac{H_l W_Q^l (W_{(K, E)}^l)^T}{\sqrt{d_k}} + B1^l_{(E)}\right) W_{(V, E)}^l + b2^l \label{eq:cl_folded}
\end{equation}

We can further fold the query projection and scaled pre-computed key matrix into a single matrix:

\begin{equation}
W_{(Q, K, E)}^l = \frac{W_Q^l (W_{(K, E)}^l)^T}{\sqrt{d_k}} \label{eq:folded_w}
\end{equation}

This yields:

\begin{equation}
    C_l = \text{ReLU}\left(H_l W_{(Q, K, E)}^l + B1^l_{(E)}\right) W_{(V, E)}^l + b2^l \label{eq:ffn_final}
\end{equation}

\subsection{Connection to Standard FFN}

The standard FFN in a Transformer block is defined as:

\begin{equation}
\text{FFN}(H_l) = \text{ReLU}(H_l W_1^l + b_1^l) W_2^l + b_2^l \label{eq:standard_ffn}
\end{equation}

Comparing this to the derived equation for \(C_l\) (Equation \ref{eq:ffn_final}), we confirm the functional equivalence \(\text{FFN}(H_l) = \text{Cross-Attention}(H_l, \text{Implicit } E)\) under the assumption of a static knowledge base \(E\). By setting:

\begin{align}
W_1^l &= W_{(Q, K, E)}^l \label{eq:mapping_w1}\\
b_1^l &= B1^l_{(E)} \label{eq:mapping_b1}\\
W_2^l &= W_{(V, E)}^l \label{eq:mapping_w2}\\
b_2^l &= b2^l \label{eq:mapping_b2}
\end{align}

the FFN becomes a specialized case of our generalized cross-attention mechanism applied to a static knowledge base \(E\), establishing functional equivalence between the two. Such equivalence directly implies that, when \(E\) is trained jointly with the model, our modular architecture is functionally equivalent to a standard Transformer. Therefore, we expect \textit{identical performance} on any task under this joint training regime. This equivalence serves as strong theoretical validation of our generalized cross-attention mechanism and our proposed modular architecture in the joint-training setting. As this implies that empirical results under joint training would simply confirm this equivalence, we defer empirical validation to future work focusing on external knowledge bases, as discussed in Section \ref{sec:discuss}.

\subsection{Implications of the Equivalence}

This mathematically proven equivalence, \(\text{FFN}(H_l) = \text{Cross-Attention}(H_l, \text{Implicit } E)\), formally establishes the connection between FFNs and the key-value memory framework~\cite{geva2020transformer}, providing a concrete mechanism for how this memory is accessed and utilized. Furthermore, it aligns with empirical observations, such as the layer-specific encoding of information found by Haider et al.~\cite{haider2024looking}. This equivalence has several important theoretical implications.

\textbf{Theoretical Justification and New Interpretation of FFNs.} This equivalence provides a strong theoretical basis for the effectiveness and interpretability of FFNs in Transformers. It reveals that they are not simply arbitrary non-linear transformations but rather perform a specific form of context-dependent knowledge retrieval from a highly compressed, distributed representation acquired during pre-training. This retrieval process involves knowledge-specific thresholding and transformation, incorporating both knowledge-specific and cross-embedding-space adjustments. This connection also provides a new lens for interpreting the folded weights (Eq.~\ref{eq:mapping_w1}-\ref{eq:mapping_b2}), which now represent a more interpretable combination of query, key, and knowledge base information.

\textbf{Distinct Requirements of Cross-Attention.} This analysis highlights the crucial differences between self-attention and cross-attention, particularly in the context of knowledge retrieval. While self-attention focuses on information exchange within a single source, cross-attention for knowledge retrieval requires mechanisms for selective retrieval and controlled transformation of information from an external source.

\textbf{Implications for Model Size.} The implicit encoding of \(E\) within the closure \(\text{FFN}(H_l) = \text{Cross-Attention}(H_l, \text{Implicit } E)\) directly explains the substantial parameter requirements of Transformers. Encoding knowledge in a distributed, compressed manner within the FFN weights requires substantial capacity. Our modular architecture, in future work with external knowledge bases, offers a potential solution to this by externalizing the knowledge base, allowing for more efficient scaling of knowledge capacity.

\textbf{Layer-Specific Views of Shared Knowledge.} Because the weight folding process involves layer-specific projection matrices (\(W_Q^l\), \(W_K^l\), \(W_V^l\)) and other layer-specific parameters and biases, each layer in a standard Transformer effectively accesses a different view of the same, implicitly encoded, shared knowledge.

\section{Discussion and Future Work}\label{sec:discuss}

This section discusses limitations, outlines future research directions, and presents practical considerations regarding computational and memory trade-offs.

\subsection{End-to-End vs. Decoupled Architectures}\label{sec:discuss:arch_comparison}

A central consideration in architectural design is the trade-off between end-to-end and decoupled (modular) approaches. End-to-end training of monolithic Transformers has proven highly effective in many tasks, offering the advantage of direct optimization for the final task objective and implicit feature learning. However, this comes at the cost of limited interpretability, adaptability, and scalability, as discussed in Section \ref{sec:challenges}. Our proposed modular architecture is theoretically equivalent to standard Transformers under the joint training regime explored in this paper (as demonstrated in Section \ref{sec:ffn}), and therefore we expect similar performance in this setting. However, its primary motivation is to address the long-term challenges of interpretability, adaptability, and scalability, particularly in scenarios requiring continuous learning and integration of rapidly evolving knowledge. We acknowledge that transitioning to external knowledge bases may introduce a performance gap, especially if knowledge representation and retrieval are not optimized. However, we argue that the potential benefits of decoupling knowledge and reasoning—including enhanced interpretability, adaptability to new information, independent scaling of knowledge and reasoning capacity, and richer interactions with external systems—outweigh this potential trade-off.

\begin{table}[t]
    \centering
    \begin{tabular}{lcc}
        \toprule
        Feature & End-to-End (Monolithic) & Decoupled (Modular) \\
        \midrule
        Model Performance & \makecell{Direct optimization \\ High} & \makecell{Equivalent (under joint training) \\ Potentially lower with external KBs} \\
        Interpretability & Limited & Enhanced \\
        Adaptability & Low, retraining/fine-tuning & High, modular updates \\
        Scalability & Limited, entangled & Enhanced, independent \\
        Inference Efficiency & Often high & Potentially lower \\
        Knowledge Representation & Implicit, distributed & Explicit, centralized (in KB) \\
        \bottomrule
    \end{tabular}
    \caption{Comparison of End-to-End and Modular Architectures with Decoupled Shared Knowledge.}
    \label{tab:arch_comparison}
\end{table}

\subsection{Limitations and Future Work}~\label{sec:discuss:future}

While this work demonstrates a functional equivalence to existing Transformers under joint training, this new perspective offers several crucial advantages motivating significant future research. It provides a rigorous theoretical foundation for understanding FFNs as performing implicit knowledge retrieval, moving beyond empirical observations. Crucially, it opens new research directions centered around external KBs and the explicit decoupling of knowledge and reasoning, enabling richer LLM-external system interactions beyond simple retrieval-augmented approaches. This focus aims to enhance adaptability, scalability, and knowledge integration.

However, this work has several limitations. Our theoretical analysis focuses on joint training, not directly addressing challenges of external, pre-existing KBs. Therefore, a primary direction for future work is the practical implementation and empirical evaluation of our modular architecture with external, pluggable KBs. This exploration raises several key research directions, encompassing the following aspects:

\textbf{External Knowledge Base Implementation and Management.} This core area of future work focuses on the practical implementation and management of external KBs within our modular architecture. It encompasses the following investigations:

\begin{itemize}
    \item \textbf{Joint Training and Retrieval.} We will investigate joint end-to-end training of the LLM and a dedicated KB embedding model \textit{X} (which generates $E$), where the LLM generates query embeddings to retrieve relevant KB entries using a differentiable Top-K approximation (e.g., smoothed softmax, straight-through estimator). This aims to optimize embedding compatibility and information integration. 
    \item \textbf{KB Storage and Management.} We will adopt external embedding storage to manage the embeddings generated by \textit{X} (i.e., $E$). This allows efficient KB updates (re-embedding, insertions, and deletions) during inference. This approach assumes sufficient training data representativeness for generalization to new KB entries, which we will evaluate. Methods for monitoring KB quality and consistency will also be investigated.
    \item \textbf{Knowledge Representation and Structure.} Throughout this paper, we have considered a simplified scenario where knowledge is represented as individual entries within the KB. However, our analysis suggests that FFNs might encode knowledge in complex, high-dimensional representations. We will therefore investigate how to represent knowledge effectively in external KBs, exploring different structures and their impact on retrieval, reasoning, and KB management. This exploration may further complicate the joint training and KB management procedures described above.
\end{itemize}

\textbf{Scaling and Efficiency.} Externalizing the knowledge base offers the potential for independent scaling of knowledge and reasoning capacity. Future work should empirically investigate the computational and memory trade-offs associated with different KB sizes, retrieval methods, and reasoning model sizes. A key research question is: \textit{How can we optimize retrieval methods, knowledge representations, and reasoning model size to achieve efficient and scalable knowledge-driven LLMs?}

\textbf{Interpretability of Retrieved Knowledge.} Understanding \textit{why} specific knowledge entries are retrieved and how they contribute to the model's output is crucial for interpretability and trustworthiness. Future work should explore methods for explaining the retrieval process, such as visualizing attention weights over retrieved knowledge entries, providing textual explanations of retrieved entries based on their content, or developing more formal methods for tracing information flow from the knowledge base to the model's predictions.

\subsection{Computational and Memory Trade-offs}

Computational efficiency is crucial. We analyze potential computational and memory trade-offs, focusing on the implications for standard Transformers given the equivalence we have shown in the joint-training setting. For simplicity, we set the dimension of the knowledge entries $d_E$ equal to the query dimension $d$. We analyze the trade-offs for different implementations of an FFN layer in a Transformer, as shown in Table~\ref{tab:comp_mem}:

\begin{table}[t]
    \centering
    \begin{tabular}{lcccc}
        \toprule
        Implementation & Standard FFN & Cross-Attention &  + Folding & + Folding + Retrieval \\
        \midrule
        Computation & $O(Nd d_{ff})$ & $O((N + d) d |E|)$ & $O(Nd|E|)$ & $O(Nd|E'|) + R$\\
        Memory & $O(d d_{ff})$ & $O(d|E|)$ & $O(d|E|)$ & $O(d|E'|)$ \\
        \bottomrule
    \end{tabular}
    \caption{Dominant Computational and Memory Complexity Terms of a Standard Transformer FFN Layer and its Equivalent Implementations using Generalized Cross-Attention. $N$ is the sequence length, $d$ is the hidden dimension for $H_l$, $d_{ff}$ is the FFN inner dimension, $|E|$ is the size of the full knowledge base, $|E'|$ is the size of the retrieved subset of the knowledge base, and $R$ represents the retrieval cost.}
    \label{tab:comp_mem}
\end{table}

\begin{itemize}
    \item \textbf{Standard FFN.} Computational complexity is $O(Nd d_{ff})$. For typical settings where $d_{ff} = 4d$ in GPT-3, this represents a significant computational burden. Memory complexity is $O(d d_{ff})$.
    \item \textbf{Cross-Attention.} A naive implementation of our generalized cross-attention involves projections and attention computation. The computational complexities of these operations are as follows:
        \begin{align*}
            \text{Query Projection} &: O(N  d  d_k) \\
            \text{Key Projection} &: O(|E|  d  d_k) \\
            \text{Value Projection} &: O(|E|  d  d) \\
            \text{Scaled Dot-Product Attention} &: O(N  d_k  |E|) \\
            \text{Multiplication by } V &: O(N   d  |E|)
        \end{align*}
    where $d_k$ represents the dimension of the keys and queries, which is usually much smaller than $d$ (e.g., 128 vs. 12288 in GPT-3). Consequently, the dominant terms are value projection and multiplication by $V$, resulting in a total complexity of $O((N + d) d |E|)$, substantially higher than the FFN when $|E| \gg d_{ff}$. Memory complexity is $O(d|E|)$. 
    \item \textbf{Cross-Attention with Folding (to full KB).} Computational complexity is reduced to $O(Nd|E|)$ due to pre-computation. Memory complexity remains $O(d|E|)$.
    \item \textbf{Cross-Attention with Folding and Retrieval (to subset $E'$).} Computational complexity becomes $O(Nd|E'|) + R$, where $R$ is the retrieval cost. If $|E'| \ll |E|$ (e.g., retrieving a few hundred to a thousand entries from a large KB), this has the potential to offer substantial computational savings. Memory complexity is reduced to $O(d|E'|)$.
\end{itemize}

This comparison highlights the fundamental trade-off between the size of the knowledge base $|E|$ and the computational cost of cross-attention, which scales linearly with $|E|$. The observation that even setting $|E| = d_{ff}$ (49152 in GPT-3 model) results in a remarkably small number of entries compared to world knowledge strongly supports our hypothesis of substantial knowledge compression within FFNs (Section~\ref{sec:discuss:future}). However, a key distinction is that in our proposed architecture, the knowledge base $E$ is \textbf{shared across all layers}, whereas in a standard Transformer, the corresponding weights within the FFNs (which implicitly encode the compressed knowledge) are \textbf{not shared}. This sharing of $E$ has important implications for parameter efficiency and knowledge consistency. It reinforces the crucial role of efficient knowledge representation (Section~\ref{sec:discuss:future}) to minimize $|E|$ and make externalization computationally feasible. Using a retrieved subset $E'$ further mitigates computational costs by focusing on relevant knowledge. While externalization introduces a retrieval cost $R$, it offers significant advantages: independent scaling of knowledge and reasoning capacity, improved adaptability, and enhanced interpretability. Future work will investigate these trade-offs, including knowledge compression techniques, the impact of retrieval methods on $R$, and the optimal size of $E'$.

\section{Related Work}\label{sec:related}

Our work draws upon and contributes to several areas of research, including Transformer architectures, knowledge retrieval, the connection between symbolic and neural AI, modular neural networks, interpretability, and generalized attention.

\textbf{Transformers, FFNs, and Knowledge.} A central challenge in integrating knowledge into Transformers is satisfying the two key requirements for a valid knowledge base (KB) discussed in Section~\ref{sec:modular:requirements}: global sharing and layer-specific views. Existing approaches struggle to reconcile these competing demands. Geva et al. \cite{geva2020transformer} proposed that FFNs function as implicit key-value memories, but this knowledge is inherently layer-local, violating the \textit{global sharing} requirement and leading to redundancy and inconsistencies. Several subsequent approaches attempt to address scaling and knowledge management, but ultimately fail to simultaneously satisfy both requirements for a valid KB. PlugLM \cite{cheng2023decouple} introduces a shared knowledge base, satisfying the \textit{global sharing} requirement, but uses \textit{identical} keys and values for all layers, thus failing to provide \textit{layer-specific views}. TokenFormer \cite{liu2024tokenformer} tokenizes layer-specific weights as a form of compressed knowledge, again violating the \textit{global sharing} requirement.

In contrast, our work directly addresses this tension by introducing a globally shared knowledge base with layer-specific transformations, achieved through a novel generalized cross-attention mechanism specifically designed for knowledge retrieval. This design uniquely satisfies both key requirements: a single, explicitly accessible knowledge base is shared across all layers (addressing parameter efficiency, knowledge consistency, and efficient management), while layer-specific transformations, implemented via distinct projection matrices and knowledge-dependent biases within our generalized cross-attention mechanism, ensure each layer accesses a unique, contextually relevant view of the knowledge (enabling more nuanced knowledge utilization). 

\textbf{Modular Neural Networks.} Modularity in neural networks has been shown to improve learning, generalization, and interpretability \cite{jacobs1991adaptive}. Previous work has explored task-specific modularity \cite{andreas2016neural,hashimoto2018joint} and parameterized Transformers \cite{yu2021parameterized}, introducing modularity at a higher level (e.g., different modules for different tasks). Our work focuses on modularity within the Transformer architecture, formalizing the FFN as a module dedicated to implicit knowledge retrieval via cross-attention. This formalization lays the groundwork for explicitly decoupling knowledge into a separate module (the globally shared KB), distinguishing our approach from previous modular neural network designs.

\textbf{Interpretability of Neural Networks.} Various techniques, such as attention visualization \cite{bahdanau2014neural}, saliency maps \cite{simonyan2013deep}, and probing tasks \cite{conneau2018cram}, aim to improve the interpretability of neural networks. While these methods provide insights into input importance, our work offers a theoretical framework for understanding the internal computations of FFNs, revealing their role as key-value memories \cite{geva2020transformer}. This understanding, in conjunction with empirical analyses like those of Haider et al. \cite{haider2024looking}, can inform more targeted interpretability methods, such as analyzing the folded weights in our derived formulation, and also provides a foundation for more interpretable architectures by explicitly separating knowledge retrieval.

\textbf{Generalized Attention and Biases.} Our work uses generalized cross-attention with knowledge-specific biases. Prior work has explored different attention mechanisms \cite{bahdanau2014neural} and biased attention \cite{shaw2018self}. We extend these by deriving the FFN as a specific biased cross-attention mechanism, demonstrating the crucial role of these biases in knowledge retrieval. The use of knowledge-specific biases, as opposed to general biases, enables finer control over retrieval and facilitates future work with external knowledge bases.

\section{Conclusion}\label{sec:cc}

We proposed a novel modular Transformer architecture with a generalized cross-attention mechanism for accessing a shared knowledge base, addressing the entanglement of knowledge and reasoning in monolithic Transformers. Our key contribution is twofold: the design of this cross-attention mechanism for effective knowledge retrieval and a theoretical analysis interpreting FFNs as a specialized case. This interpretation reveals FFNs perform implicit knowledge retrieval and motivates future research exploring external knowledge bases to enhance adaptability, scalability, and richer LLM-external system interactions beyond simple retrieval-augmentation. This modular design offers a promising avenue for more interpretable and scalable knowledge-driven AI.

\section*{Acknowledgments}
The authors gratefully acknowledge Xiyou Guo for his valuable contribution in identifying and disproving initial hypotheses through mathematical analysis. 

\bibliographystyle{plain} 
\bibliography{refs} 

\end{document}